\pdfoutput=1

\documentclass[11pt]{article}

\usepackage{acl}

\usepackage{times}
\usepackage{latexsym}

\usepackage{booktabs}
\usepackage{multirow}
\usepackage{xcolor}
\usepackage{colortbl}
\usepackage{array} 
\usepackage{amsmath}

\newcolumntype{C}{>{\centering\arraybackslash}p{0.07\textwidth}}
\usepackage[T1]{fontenc}
\usepackage[utf8]{inputenc}

\usepackage{microtype}
\DeclareMathOperator*{\argmax}{arg\,max}
\usepackage{inconsolata}

\usepackage{graphicx}
\title{Wi-Chat: Large Language Model Powered Wi-Fi Sensing}

\author{
{Haopeng Zhang$\dag$\thanks{These authors contributed equally to this work.}, Yili Ren$\ddagger$\footnotemark[1], Haohan Yuan$\dag$, Jingzhe Zhang$\ddagger$, Yitong Shen$\ddagger$} \\
ALOHA Lab, University of Hawaii at Manoa$\dag$, University of South Florida$\ddagger$ \\
\{haopengz, haohany\}@hawaii.edu\\
\{yiliren, jingzhe, shen202\}@usf.edu\\}

\begin{document}
\maketitle
\begin{abstract}
Recent advancements in Large Language Models (LLMs) have demonstrated remarkable capabilities across diverse tasks. However, their potential to integrate physical model knowledge for real-world signal interpretation remains largely unexplored. In this work, we introduce Wi-Chat, the first LLM-powered Wi-Fi-based human activity recognition system. We demonstrate that LLMs can process raw Wi-Fi signals and infer human activities by incorporating Wi-Fi sensing principles into prompts. Our approach leverages physical model insights to guide LLMs in interpreting Channel State Information (CSI) data without traditional signal processing techniques. Through experiments on real-world Wi-Fi datasets, we show that LLMs exhibit strong reasoning capabilities, achieving zero-shot activity recognition. These findings highlight a new paradigm for Wi-Fi sensing, expanding LLM applications beyond conventional language tasks and enhancing the accessibility of wireless sensing for real-world deployments.
\end{abstract}

\section{Introduction}
\vspace{-3mm}


Large Language Models (LLMs) have achieved significant advancements with ChatGPT representing a particularly revolutionary milestone. Their ability to generate human-like conversations in a zero-shot manner signifies a transformative shift in human-AI interactions. Moreover, the latest LLMs demonstrated remarkable reasoning capabilities and exceptional generalization skills. However, their reliance on training with collections of textual content from the Internet leaves them considerably distant from achieving a profound understanding of the physical world. Meanwhile, the ubiquitous Wi-Fi devices and the extensive coverage of Wi-Fi networks present an opportunity to expand Wi-Fi capabilities beyond communication, particularly in sensing the physical world~\cite{tan2022commodity}. As Wi-Fi signals traverse the physical environment, they interact with surrounding people and objects, causing reflection, diffraction, scattering, etc. Consequently, the received signals can carry a substantial amount of information about both people and the environment. Conventional Wi-Fi-based sensing systems can achieve various sensing tasks in the physical environment, such as human activity recognition and localization~\cite{ma2019wifi}. However, these systems typically rely on complex signal processing techniques and the labor-intensive training of machine learning or deep learning models.
This raises a fundamental and compelling question: \emph{can we integrate LLMs with Wi-Fi sensing to comprehend the physical world without complex signal processing and in a zero-shot manner?}


We investigate this question by exploring the capabilities of LLMs to understand raw Wi-Fi signals and incorporating physical model guidance based on Wi-Fi sensing. Specifically, we introduce Wi-Chat, an LLM-powered Wi-Fi sensing system for human activity recognition. Unlike existing LLMs that primarily analyze traditional textual and visual data, Wi-Chat can understand Wi-Fi signals which are real-world projections of the physical environment. We demonstrate that LLMs, having been trained on extensive human knowledge, when integrated with the physical models of Wi-Fi sensing, can be directly leveraged for Wi-Fi signal analysis. This approach can derive deep insights that traditionally require complex signal processing and machine learning or deep learning models trained on large volumes of labeled data.

Wi-Chat directly feeds textual or visual representations of raw Wi-Fi signals into well-known LLMs such as ChatGPT, DeepSeek, and LLama for human activity recognition. Additionally, we incorporate physical models of Wi-Fi sensing into LLMs via prompting, enabling a deep understanding of human activities in the physical world through Wi-Fi signals.
We conduct experiments using a self-collected human activity dataset. The benchmark systems for comparison include both conventional Wi-Fi-based human activity recognition systems and fundamental machine learning models. Our evaluations demonstrate that LLMs can perform zero-shot human activity recognition directly from raw Wi-Fi signals, achieving the best accuracy of $90\%$ on unseen data. The key contributions of our work include:

\begin{itemize}
  \item We propose Wi-Chat, the first LLM-powered Wi-Fi-based human activity recognition system, which integrates the reasoning capabilities of LLMs with the sensing capabilities of wireless signals to interpret human activities.
  \vspace{-3mm}
  \item We incorporate Wi-Fi sensing into prompts, providing physical model guidance to LLMs.
  \vspace{-3mm}
  \item The experiments show that Wi-Chat achieves promising performance in a zero-shot setting without complex signal processing.
\end{itemize}


\begin{figure*}[ht]
    \centering
    \includegraphics[width=0.95\textwidth]{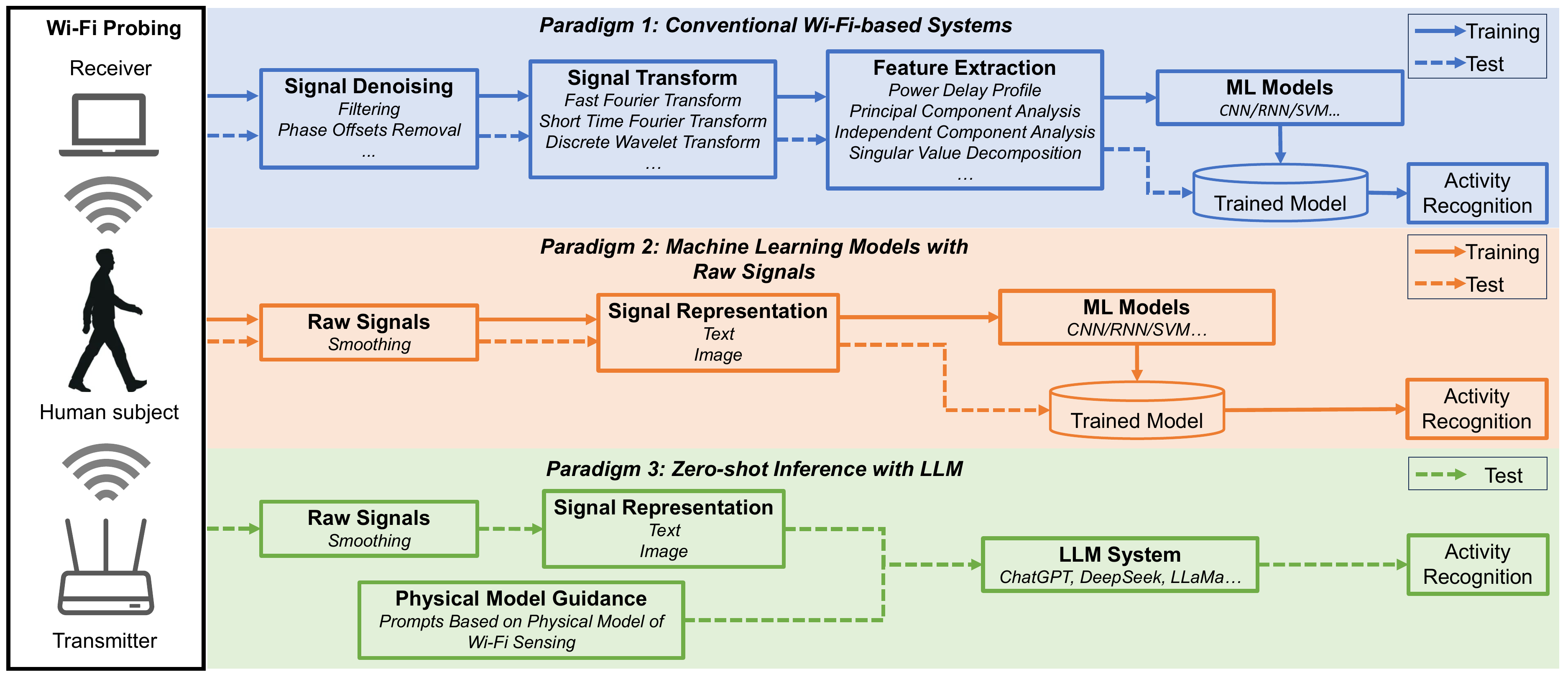}
    \vspace{-3mm}
    \caption{The overall computing pipeline of three different paradigms for Wi-Fi-based human activity recognition.}
    \vspace{-4mm}
    \label{systemflow}
\end{figure*}

\section{Related Work}

\textbf{Wi-Fi Sensing.} Wi-Fi sensing has been extensively explored for various applications~\cite{qian2017inferring, xie2019md, ren2020liquid, jiang2020towards, ren2023person, li2024spacebeat, wang2024multi}, particularly in human activity recognition~\cite{wang2015understanding, zhang2017toward, wang2014eyes}, due to its non-contact nature and low cost.
For example, E-eyes~\cite{wang2014eyes} was the first work to leverage Wi-Fi signals for recognizing daily human activities. \citet{zhang2017toward} theoretically analyzed the perception capabilities of Wi-Fi signals and introduced a Fresnel zone model for human activity sensing. WiG~\cite{he2015wig} utilizes support vector machines to extract activity-related features from Wi-Fi. CARM~\cite{wang2015understanding} employs a hidden Markov model to extract temporal features from Wi-Fi for activity recognition. Additionally, \citet{yang2019learning} applied convolutional and recurrent neural networks to extract distinguishing features from Wi-Fi signals for human activity recognition.
While these conventional Wi-Fi sensing systems achieve strong performance, they typically rely on multi-stage signal processing techniques and require large amounts of data to train deep learning or machine learning models.

\noindent
\textbf{Large Language Model Applications.} Recently, the emergence of LLMs has revolutionized both academic NLP research and industrial
products due to their remarkable ability to understand, analyze, and generate texts with vast
amounts of pre-trained knowledge~\cite{zhao2023survey,zhang2023summit,liu2023pre,zhang2024systematic}. By leveraging extensive corpora of text data, LLMs can capture
complex linguistic patterns, semantic relationships, and contextual cues, enabling them to produce
high-quality responses. LLMs have also been applied beyond the scope of NLP, as a powerful tool that has
propelled the field of healthcare, legal and finance~\cite{chen2024survey,yuan2024structure,he2025survey}. 
More recently, researchers have explored the application of LLMs in sensing-related domains. The concept of Penetrative AI~\cite{xu2024penetrative} has been introduced to integrate LLMs with the physical world, enabling the analysis of sensor data through LLMs. Additionally, HARGPT~\cite{ji2024hargpt} and LLMTrack~\cite{yang2024you} have demonstrated the potential of LLMs in recognizing human activities and tracking robotic movements by analyzing Inertial Measurement Unit (IMU) data. These advancements highlight the growing potential of bridging LLMs with physical world sensing applications.


\section{Preliminary}

\subsection{Wi-Fi Sensing} \label{Preliminary}

In recent years, Wi-Fi sensing has garnered significant attention due to the widespread use of Wi-Fi devices and their ability to sense people and surrounding environments. In an indoor setting, Wi-Fi signals propagate through both direct (i.e., line-of-sight (LoS)) and reflected paths, bouncing off objects and humans, before reaching the receiver. When the sensing target is a person, changes in the signal can be leveraged to infer locations, activities, and even vital signs of the person, enabling contactless human sensing. Moreover, existing Wi-Fi infrastructure can be reused for sensing, allowing for seamless integration into an LLM-powered smart environment.

To capture changes in Wi-Fi signals caused by a target (e.g., a person), we utilize Channel State Information (CSI), which characterizes how signals are altered as they propagate through physical space. Specifically, Wi-Fi signals travel from a transmitter to a receiver through multiple paths, including a direct path (i.e., LoS propagation) and numerous reflected paths from objects such as walls, furniture, and the person. CSI represents the superposition of signals from all these paths and describes this multipath propagation, which can be expressed as: $H(f, t) = \sum^{N}_{i=1}a_{i}e^{-j2\pi \frac{d_{i}(t)}{\lambda}}$,
where $a_{i}$ is the complex attenuation and $d_{i}(t)$ is the length of the $i^{th}$ path, $N$ is the number of paths, $\lambda$ is the wavelength, and $f$ is the signal frequency.

The CSI can be further decomposed into static and dynamic components. The static component consists of the LoS signals and reflections from stationary objects in the environment. In contrast, the dynamic component arises from reflections caused by the moving target. For simplicity, we assume that there is only a single signal reflection from the target. Thus, the CSI can be denoted as: 
\vspace{-2mm}
\begin{equation} \label{eq2}
\begin{split}
H(f, t) &= H_{s}(f, t) + H_{d}(f, t) \\
&= H_{s}(f, t) + a(f,t)e^{-j2\pi \frac{d(t)}{\lambda}},
\end{split}
\end{equation}
where $H_{s}(f, t)$ is the static component, $a(f,t)$, $e^{-j2\pi \frac{d(t)}{\lambda}}$, and $d(t)$ are the complex attenuation, phase shift and path length of dynamic component $H_{d}(f, t)$, respectively.

\subsection{Wi-Fi-based Human Activity Recognition} \label{systems}
The overall system flow for Wi-Fi-based human activity recognition is shown in Figure~\ref{systemflow}. In all systems, a Wi-Fi transmitter emits signals that are received by Wi-Fi receivers to probe human activities. The system processes Wi-Fi CSI measurements, which can be extracted from the network interface controllers of Wi-Fi devices.
In this work, we examine three distinct types of paradigms:

\textbf{Paradigm 1: Conventional Wi-Fi-based Systems.} Following Wi-Fi probing, raw Wi-Fi signals undergo processing in a signal-denoising module to reduce noise. This process may involve phase offset removal techniques to calibrate the signal phase~\cite{guo2017wifi, kotaru2015spotfi} and various filtering methods to eliminate outliers~\cite{ali2017recognizing}.
Next, signal transformation methods are applied for time-frequency analysis of the denoised Wi-Fi signals. Common techniques include Fast Fourier Transform (FFT), Short-Time Fourier Transform (STFT), and Discrete Wavelet Transform (DWT).
Next, a feature extraction step extracts relevant features from the preprocessed Wi-Fi signals. For instance, propagation distance features can be derived using the Power Delay Profile (PDP)~\cite{xie2015precise}. Additionally, Principal Component Analysis (PCA), Independent Component Analysis (ICA), and Singular Value Decomposition (SVD) are widely employed for feature extraction, signal separation, and dimensionality reduction.
Once features are extracted, deep learning or machine learning models are trained to learn a mapping function between Wi-Fi signals and corresponding ground truth activity labels. After training, the model can perform human activity recognition on newly captured Wi-Fi signals.

\textbf{Paradigm 2: Machine Learning Models with Raw Signals.}
An alternative approach is directly inputting raw Wi-Fi signals into machine learning models, including Convolutional Neural Networks (CNNs), Recurrent Neural Networks (RNNs), and Support Vector Machines (SVMs).
In this approach, we first apply a simple signal smoothing. The signals are then converted into textual or visual representations before being fed into the models. Similar to conventional systems, these models still require extensive manual labeling and training.

\textbf{Paradigm 3: LLM-based Zero-shot Inference System.} In Wi-Chat, the input data consists of textual or visual representations of raw Wi-Fi signals, processed with simple signal smoothing. The system aims to recognize human activities based on these signals.
To achieve this, we explicitly instruct LLMs to determine a person's activity by analyzing the Wi-Fi signals. The underlying idea is that different human activities induce distinct patterns in Wi-Fi signals. By integrating physical models of Wi-Fi sensing into prompts, we provide physical model guidance to LLMs to interpret the signals.
Unlike conventional Wi-Fi-based systems or machine learning models, Wi-Chat eliminates the need for complex signal processing and labor-intensive model training, offering a more efficient and scalable approach to activity recognition.

\begin{figure}[t]
    \centering
    \includegraphics[width=0.42\textwidth]{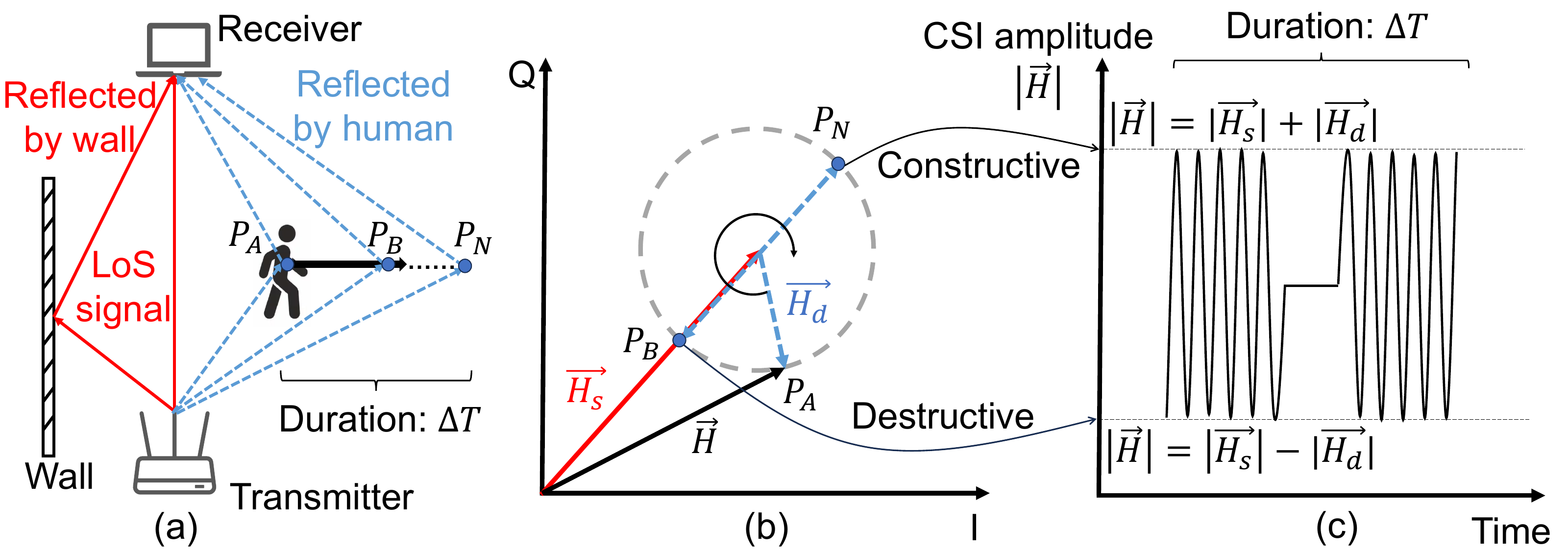}
    \vspace{-3mm}
    \caption{Modeling the human walking scenario.}
    \vspace{-4mm}
    \label{csiwalk}
\end{figure}

\begin{figure}[t]
    \centering
    \includegraphics[width=0.42\textwidth]{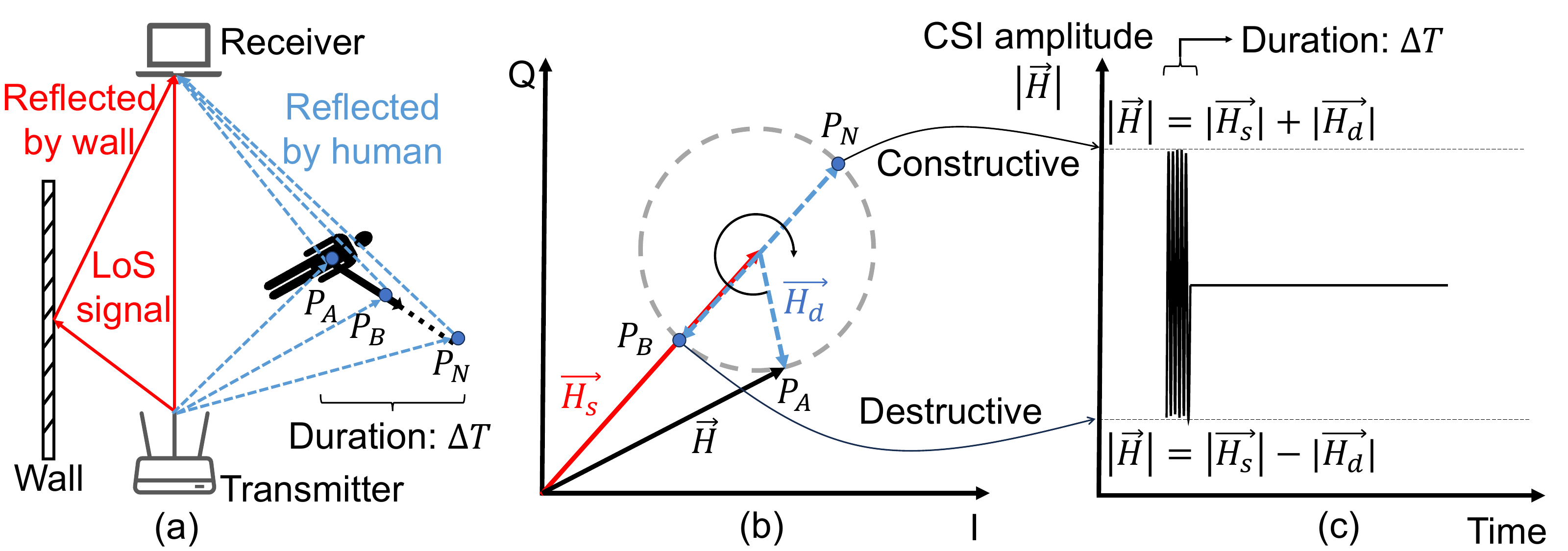}
    \vspace{-3mm}
    \caption{Modeling the human falling scenario.}
    \vspace{-7mm}
    \label{csifall}
\end{figure}

\begin{figure}[t]
    \centering
    \includegraphics[width=0.42\textwidth]{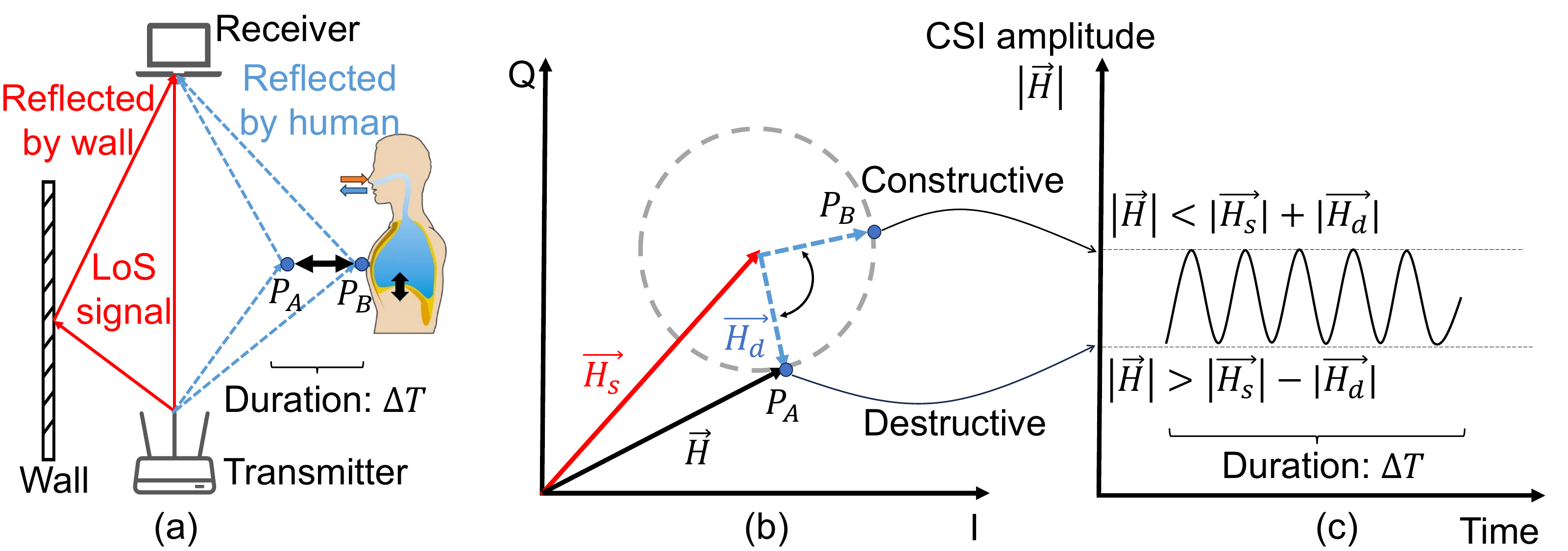}
    \vspace{-3mm}
    \caption{Modeling the human breathing scenario.}
    \vspace{-4mm}
    \label{csibreath}
\end{figure}

\begin{figure}[t]
    \centering
    \includegraphics[width=0.42\textwidth]{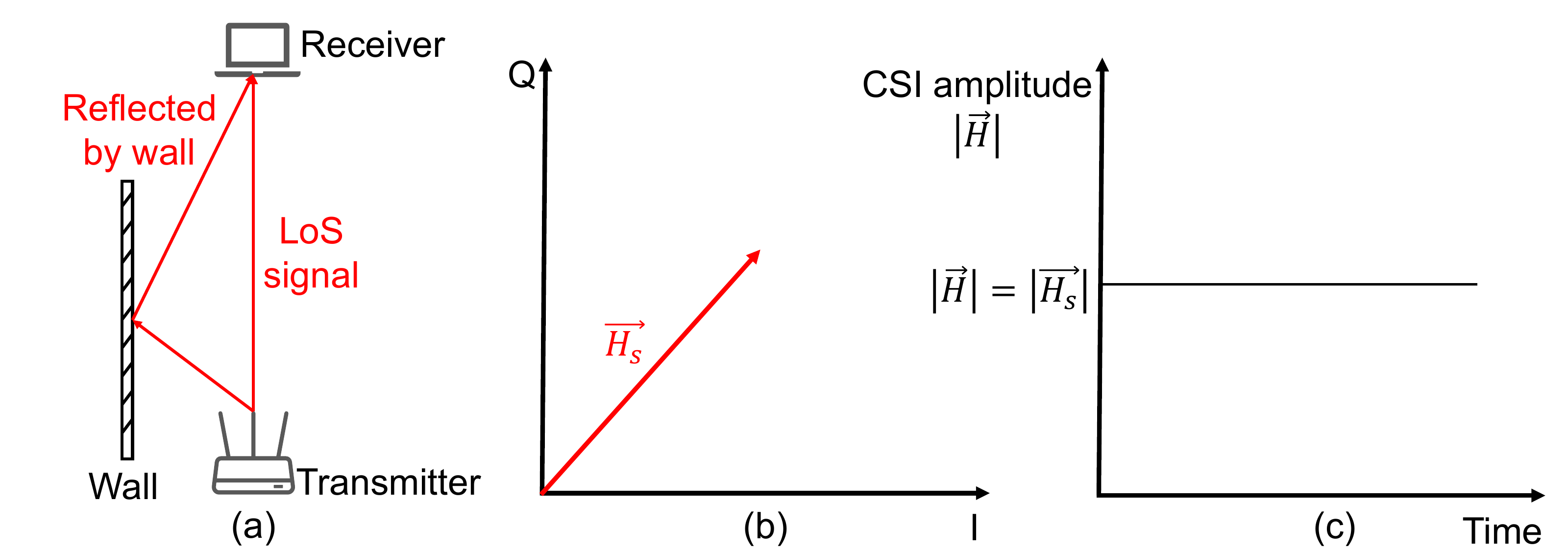}
    \vspace{-3mm}
    \caption{Modeling the no-event scenario.}
    \vspace{-7mm}
    \label{csino}
\end{figure}

\begin{figure*}[h]
    \centering
    \includegraphics[width=0.9\textwidth]{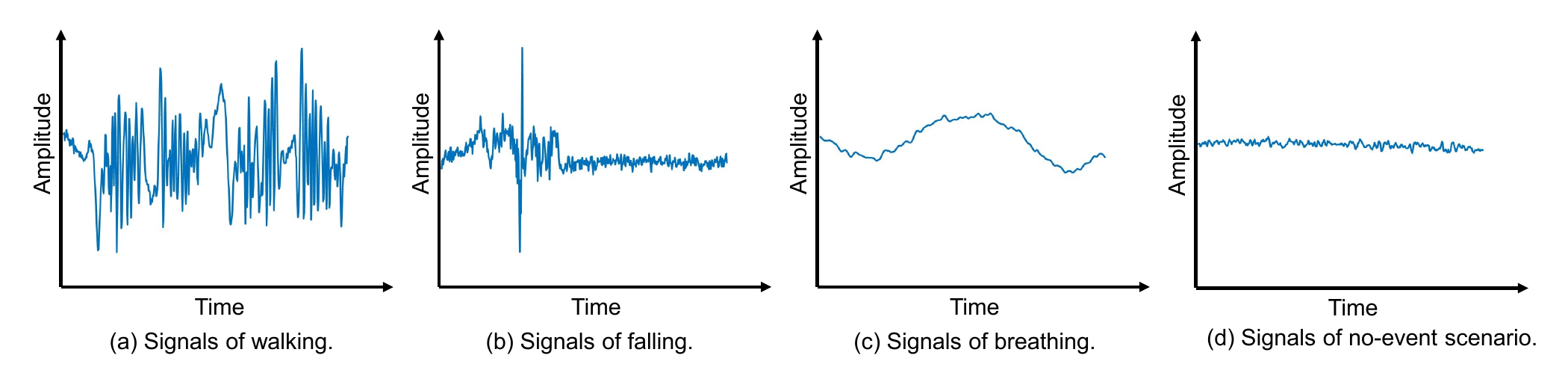}
    \caption{Real signals of different human activities.}
    \label{signals}
\end{figure*}

\section{Method}
\subsection{Wi-Fi Physical Model Knowledge}
\label{phisical}
In this subsection, we construct the physical models of Wi-Fi sensing in terms of human walking, falling, breathing, and no-event scenarios. Then we derive the LLM prompting guidance accordingly.


\textbf{Modeling the Human Walking Scenario.} As shown in Figure~\ref{csiwalk}(a), a person is walking and the Wi-Fi transmitter and receiver are placed at fixed locations. The Wi-Fi transmitter emits Wi-Fi signals which propagate through the multipath environments. As described in Section~\ref{Preliminary}, the static components (i.e., solid red lines) include the LoS signal and the signals reflected off the wall. The dynamic components (i.e., dotted blue lines) consist of the signals reflected off the human body. Assuming the person moves from location $P_A$ to location $P_N$, this movement results in a change in the path length of the dynamic component. The duration of this movement is denoted as $\Delta T$. 



We can plot the signals based on the in-phase (I) and quadrature (Q) components~\cite{wang2015understanding} in the IQ plane, as illustrated in Figure~\ref{csiwalk}(b). Theoretically, the static component vector ($\vec{H_s}$) remains fixed, while the dynamic component vector ($\vec{H_d}$) can change and rotate. The overall CSI $\vec{H}$ is the sum of vectors $\vec{H_s}$ and $\vec{H_d}$. When the dynamic and static component vectors on the IQ plane are aligned in the same direction (e.g., at $P_{N}$), they add constructively, resulting in the maximum CSI amplitude (i.e., $|\vec{H}|=|\vec{H_s}|+|\vec{H_d}|$). Conversely, when they are in opposite directions (e.g., at $P_{B}$), they add destructively, minimizing the CSI amplitude (i.e., $|\vec{H}|=|\vec{H_s}|-|\vec{H_d}|$). Note that when the path length of the dynamic component changes by one wavelength (e.g., one wavelength is about 6 $cm$ for 5 GHz Wi-Fi), its phase rotates by $2\pi$~\cite{ren2021winect, ren2022gopose} as indicated in Equation~\ref{eq2}. Since human walking is a large-scale activity, each step can cause changes of many wavelengths in the propagation path, resulting in multiple phase rotations in the dynamic components. This leads to multiple peaks ($|H_s|+|H_d|$) and troughs ($|H_s|-|H_d|$) for CSI amplitude ($|\vec{H}|$) as shown in Figure~\ref{csiwalk}(c). In this work, we focus exclusively on the CSI amplitude. Additionally, we note that human walking is a continuous activity with a duration $\Delta T$ that can range from a few seconds to several minutes.

Thus, we formulate the LLM prompting guidance for human walking as follows: 
\emph{``Walking is a large-scale activity that induces significant changes in the Wi-Fi CSI amplitude over time, characterized by the presence of numerous peaks and troughs.''}





\textbf{Modeling the Human Falling Scenario.} Similar to walking, the falling scenario is also a large-scale activity, where the person moves from $P_A$ to $P_N$, as depicted in Figure~\ref{csifall}(a). This movement can cause the overall CSI amplitude ($|\vec{H}|$) to reach both maximum and minimum values, corresponding to $|\vec{H_s}| + |\vec{H_d}|$ and $|\vec{H_s}| - |\vec{H_d}|$, respectively, as shown in Figures~\ref{csifall}(b) and (c). However, the duration $\Delta T$ of a fall can be very short (e.g., about 0.5 seconds)~\cite{choi2015kinematic}. As a result, the peaks and troughs caused by the fall are concentrated within a brief period. After signal smoothing, these rapid fluctuations can be considered as one significant peak/trough. Following the fall, the person may lose consciousness or remain motionless, leading to a static period after the fall.

Therefore, we can formulate the LLM prompting guidance for human falling as follows:
\emph{``Falling is a large-scale and sudden activity that induces a single significant peak/trough in the Wi-Fi CSI amplitude, followed by a relatively stable period.''}

\textbf{Modeling the Human Breathing Scenario.} Human breathing is a small-scale activity, as the typical range of chest expansion and contraction during a breath is only a few centimeters (from $P_A$ to $P_B$, as shown in Figure~\ref{csibreath}(a)). This movement leads to dynamic path length changes that are typically very small. Thus, the overall CSI amplitude may not reach both its maximum and minimum values, meaning that $(|\vec{H_s}| - |\vec{H_d}|)<|\vec{H}|<(|\vec{H_s}| + |\vec{H_d}|)$, as illustrated in Figures~\ref{csibreath}(b) and (c). Furthermore, breathing is a continuous and smooth activity, meaning its duration ($\Delta T$) is long. 


We formulate the LLM prompting guidance for human breathing as follows:
\emph{``Breathing is a small-scale and smooth activity that causes slow and gradual changes in Wi-Fi CSI amplitude over time, with a moderate variation range.''}

\textbf{Modeling the No-event Scenario.} In this scenario, only static signal components exist, such as the LoS signals and the signals reflected by stationary objects, as depicted in Figures~\ref{csino}(a) and (b). Since no movement occurs, no dynamic component is introduced into the Wi-Fi signal propagation. As a result, the overall CSI amplitude is determined solely by the amplitude of these static signal components and remains nearly constant over time (i.e., $(|\vec{H}|=|\vec{H_s}|$), as illustrated in Figure~\ref{csino}(c).

We formulate the LLM prompting guidance for the absence of motion as follows: 
\emph{``For the no-event scenario, the time-series CSI amplitude remains stable, meaning the variation range is very small.''}

We further illustrate examples of real Wi-Fi signals corresponding to different human activities in Figure~\ref{signals}. These signal patterns align well with our physical model for Wi-Fi sensing, validating its effectiveness in characterizing human activities through Wi-Fi CSI.
By leveraging these physical insights, our derived prompts offer explicit guidance to LLMs, enabling them to interpret Wi-Fi signals based on Wi-Fi sensing principles. This allows for accurate recognition of human activities in a zero-shot manner.


\vspace{-2mm}
\subsection{Wi-Chat: LLMs for Wi-Fi-Based Activity Recognition}

This section outlines different prompting strategies for leveraging LLMs in Wi-Fi-based human activity recognition. We aim to explore how LLMs can interpret Wi-Fi signals and improve activity classification without extensive model training or complex signal processing.

\textbf{Base.} 
For the base setting, we provide the LLM with raw CSI amplitude data, represented as a time series, and prompt it to recognize human activity labels, represented as $\hat{a}=\argmax_a p_{M}(a\mid s)$. Additionally, incorporating the physical model of Wi-Fi sensing, as described in Section~\ref{phisical}, as \textbf{domain knowledge} may enhance the LLM's interpretability and performance.

\textbf{In-context Learning.}
Recent studies have demonstrated that LLMs exhibit strong few-shot learning capabilities across various tasks, a phenomenon known as in-context learning (ICL)~\cite{brown2020language}. By learning from these exemplars within a single inference session, the model can recognize patterns in the signals and improve its classification accuracy without additional fine-tuning. The standard ICL prompts a large language model, $M$, with a set of $k$ exemplar and predicts a activity $\hat{a}$ for the Wi-Fi signal by:
\vspace{-3mm}
\begin{equation}
    \hat{a}=\argmax_a p_{M}(a\mid s,\{(s^1,a^1)...(s^k,a^k)\}).
\end{equation}

\textbf{CoT.} Beyond simple input-output mappings, incorporating chain-of-thought (CoT) reasoning into prompts can further enhance the model's interpretability~\cite{nye2021show, wei2022chain}. By including explicit intermediate steps, CoT prompting helps the model better capture the relationships between signal variations and human activities. It could be represented as:
\vspace{-3mm}
\begin{equation}
   \hat{a}=\argmax_a p_{M}(a\mid s,C),
\end{equation}
where $C=\{(s^1,e^1,a^1)...(s^k,e^k,a^k)\}$ is the set of input-explanation-output triplets in prompts. 

\textbf{Multi-modal.}
Since raw Wi-Fi signals in numerical form can be difficult for LLMs to interpret, we extend our approach by incorporating visual representations. Specifically, we generate signal plots and present them as additional inputs, allowing the model to process both textual and visual information. This multi-modal strategy leverages the LLM's capability to analyze images, potentially improving activity recognition by making signal variations more interpretable.
\vspace{-3mm}
\begin{equation}
   \hat{a}=\argmax_a p_{M}(a\mid s,v),
\end{equation}
where $v$ represents the visual representation of the signal. By incorporating these plots, we aim to improve the interpretability of the Wi-Fi signals for LLMs, enabling more accurate activity recognition.

By exploring these prompting strategies, we aim to assess the feasibility of LLMs for Wi-Fi-based activity recognition and understand how different types of input representations influence their performance. 
\section{Experiment}
\subsection{Wi-Chat Dataset}

\begin{table}[ht]
    \centering
    \small
    \renewcommand{\arraystretch}{1.1} 
    \setlength{\tabcolsep}{6pt} 
    \begin{tabular}{llcccc}
        \toprule
        \multicolumn{2}{l}{\textbf{}} & \textbf{B} & \textbf{W} & \textbf{F} & \textbf{N} \\ 
        \midrule
        \multirow{2}{*}{\textbf{Image}} & \# Training & 86 & 63 & 63 & 63 \\
                                         & \# Test     & 37 & 27 & 27 & 27 \\ 
        \midrule
        \multirow{2}{*}{\textbf{Text}}  & \# Training & 86 & 63 & 63 & 63 \\
                                         & \# Test     & 37 & 27 & 27 & 27 \\ 
        \bottomrule
    \end{tabular}
    \caption{Statistics of the experimental dataset. B stands for Breathing, W stands for Walking, F stands for falling and N stands for No event.}
        \label{tab:data_stats}
\end{table}

We conducted experiments using a self-collected Wi-Fi CSI dataset, leveraging commodity Wi-Fi devices, specifically Dell LATITUDE laptops, as both the Wi-Fi transmitter and receivers for data collection. Each Wi-Fi transmitter and receiver is equipped with three antennas. The Wi-Fi channel operated at 5.32 GHz with a bandwidth of 40 MHz, and the packet transmission rate was set to 1000 packets per second. We utilized the Linux 802.11 CSI tool~\cite{halperin2011tool} to extract CSI data from 30 OFDM subcarriers per packet. The dataset comprises over 1,965,000 Wi-Fi CSI packets collected from participants with varying heights, weights, and ages. These packets were segmented into 393 segments, each lasting 5 seconds, during which participants performed one of four activities: walking, falling, breathing, or no event (i.e., an empty environment). The collected data were then converted into both image and text representations, as detailed in Table~\ref{tab:data_stats}. Data collection was conducted across three real-world environments, a bedroom, a kitchen, and a living room, over a two-month period. The study was reviewed and approved by the IRB of the authors' institution.


\vspace{-1mm}
\subsection{Baselines}
\vspace{-2mm}
We compare Wi-Chat with the following systems: 

\textbf{Conventional Wi-Fi-based Systems.} These systems follow a multi-step pipeline, including signal denoising, signal transformation, feature extraction, and model construction, as described in Section~\ref{systems}. Specifically, we reproduce two well-known systems:
\textit{1) CARM}~\cite{wang2015understanding}: It utilizes a PCA-based method for signal denoising, applies DWT for feature extraction, and employs a Hidden Markov Model for activity recognition.
\textit{2) E-eyes}~\cite{wang2014eyes}: This system first removes data outliers using a low-pass filter and then builds activity classifiers using Earth Mover's Distance.

\textbf{Machine Learning Models with Raw Signals.} 
We evaluate the performance of machine learning models, including \textit{3) CNN}, \textit{4) RNN}, and \textit{5) SVM}. These models take textual or visual representations of raw Wi-Fi signals as input and are trained in a supervised manner using labeled datasets.

\subsection{Experimental Settings}

For LLMs, we first apply signal smoothing using the Savitzky-Golay filter~\cite{schafer2011savitzky} and then convert the signals into textual or visual representations. When experimenting with the few-shot setting, we pick 4 examples, including one example from each label class. The prompts used for experiments are presented in Appendix~\ref{sec:prompt}.


For all supervised baselines, we randomly split the dataset into $70\%$ for training and $30\%$ for testing to ensure fair evaluation. Prior to model training, we apply the same signal smoothing techniques and convert the signals into textual and visual representations for consistency across methods. The CNN and RNN models are trained on an NVIDIA GeForce RTX 4090 GPU with a learning rate of 0.001 using the Adam optimizer, with a maximum of 30 epochs and a batch size of 32. For the SVM model, we use the Radial Basis Function as the kernel type. For CARM and E-eyes, we follow their original signal processing pipelines, including denoising, feature extraction, and model construction as described in their respective works. 
Zero-shot evaluations for CNN, RNN, SVM, and conventional systems are also conducted using the same approaches, but with untrained models to test their zero-shot performance. Performance evaluation is conducted using standard classification metrics, including accuracy, precision, recall, and F1 score, to assess the ability of each system to recognize human activities from Wi-Fi CSI data.

\vspace{-2mm}
\begin{table}[tb]
    \centering
    \small
    \begin{tabular}{lcc}
        \toprule
        \textbf{Category} & \textbf{Model} & \textbf{Accuracy} \\
        \midrule
        Zero-shot & GPT-4o & 0.62 \\
        4-shot & GPT-4o & 0.77 \\
        Zero-shot (V) & GPT-4o-mini + CoT & 0.90 \\
        Supervised (V) & CNN & 0.98 \\
        Supervised & E-eyes & 1.00 \\
        \bottomrule
    \end{tabular}
    \caption{Best performing models in each category. V stands for vision input.}
    \label{summary}
\end{table}

\subsection{Results}
\vspace{-2mm}

\textbf{Overall Results.} Table~\ref{summary} presents the best-performing models across different method categories. In the zero-shot category, the best model GPT-4o model achieved an accuracy of 0.62, demonstrating its ability to generalize effectively without task-specific examples. In the 4-shot category, GPT-4o is still the best model, exhibiting substantial improvement, attaining an accuracy of 0.77. This result highlights the effectiveness of in-context learning, where additional prompt examples help refine model predictions.

In the vision models category, the GPT-4o-mini with CoT demonstrated the strongest performance with an accuracy of 0.90. This result indicates the model's capacity to integrate visual and textual reasoning through CoT prompting, which likely aids in complex decision-making.

For supervised learning, the E-eyes (with complex signal processing techniques) outperformed all other models. This result is expected, as supervised models are explicitly trained on labeled data, allowing them to learn precise decision boundaries. However, despite the high accuracy, supervised methods typically require extensive labeled datasets, which may not always be feasible in real-world applications.





\begin{table}[th]
    \centering
    \small
    \begin{tabular}{lcc}
        \toprule
        \textbf{Method} & \textbf{Accuracy} & \textbf{F1-score} \\
        \midrule
        \multicolumn{3}{c}{\textbf{Zero Shot}} \\
         \midrule
         E-eyes & 0.26 & 0.26 \\
         CARM & 0.24 & 0.24 \\
         SVM & 0.27 & 0.27 \\
         CNN & 0.23 & 0.23 \\
         RNN & 0.26 & 0.26 \\
        Vision SVM & 0.26 & 0.25 \\
        Vision CNN & 0.26 & 0.26 \\
        Vision RNN & 0.28 & 0.28 \\
        GPT-4o & 0.62 & 0.42 \\
        GPT-4o+ ICL & 0.77 & 0.73 \\
        GPT-4o-mini + COT & \textbf{0.90} & \textbf{0.90} \\
        \midrule
        \multicolumn{3}{c}{\textbf{Supervised}} \\
         \midrule
        Vision CNN & 0.98 & 0.98 \\
        CARM & 0.98 & 0.98 \\
        E-eyes & 1.00 & 1.00 \\
        \bottomrule
    \end{tabular}
    
    \caption{Performance comparison of different methods under the zero-shot and supervised settings.}
    \label{present}
\end{table}

Overall, the results indicate that LLMs exhibit strong performance in zero-shot and few-shot settings for the task of Wi-Fi-based human activity recognition, making them valuable for scenarios with limited annotated data. Additionally, the impressive accuracy of the vision-language model suggests promising directions for integrating multimodal learning into the task.

\textbf{Methods Comparison.}
Table~\ref{present} summarizes the classification results across different methods. In the zero-shot setting, traditional machine learning models such as SVM, CNN, and RNN demonstrate relatively low classification performance. LLM models GPT-4o achieved an accuracy of 0.61, significantly outperforming traditional machine learning models. The ICL approach further improved performance, demonstrating the benefits of in-context learning.

Furthermore, GPT-4o-mini combined with CoT reasoning achieves the highest accuracy among zero-shot methods at 0.90, demonstrating the effectiveness of advanced prompting techniques in enhancing LLM-based classification performance. Notably, this performance is already comparable to conventional Wi-Fi activity recognition systems and machine learning models trained in supervised settings. These results reaffirm that while supervised models achieve superior performance with labeled data, LLMs exhibit strong generalization capabilities, particularly when leveraging few-shot learning and vision-language integration.

\subsection{Analysis}

\begin{figure}[tb]
    \centering
    \includegraphics[width=0.49\textwidth]{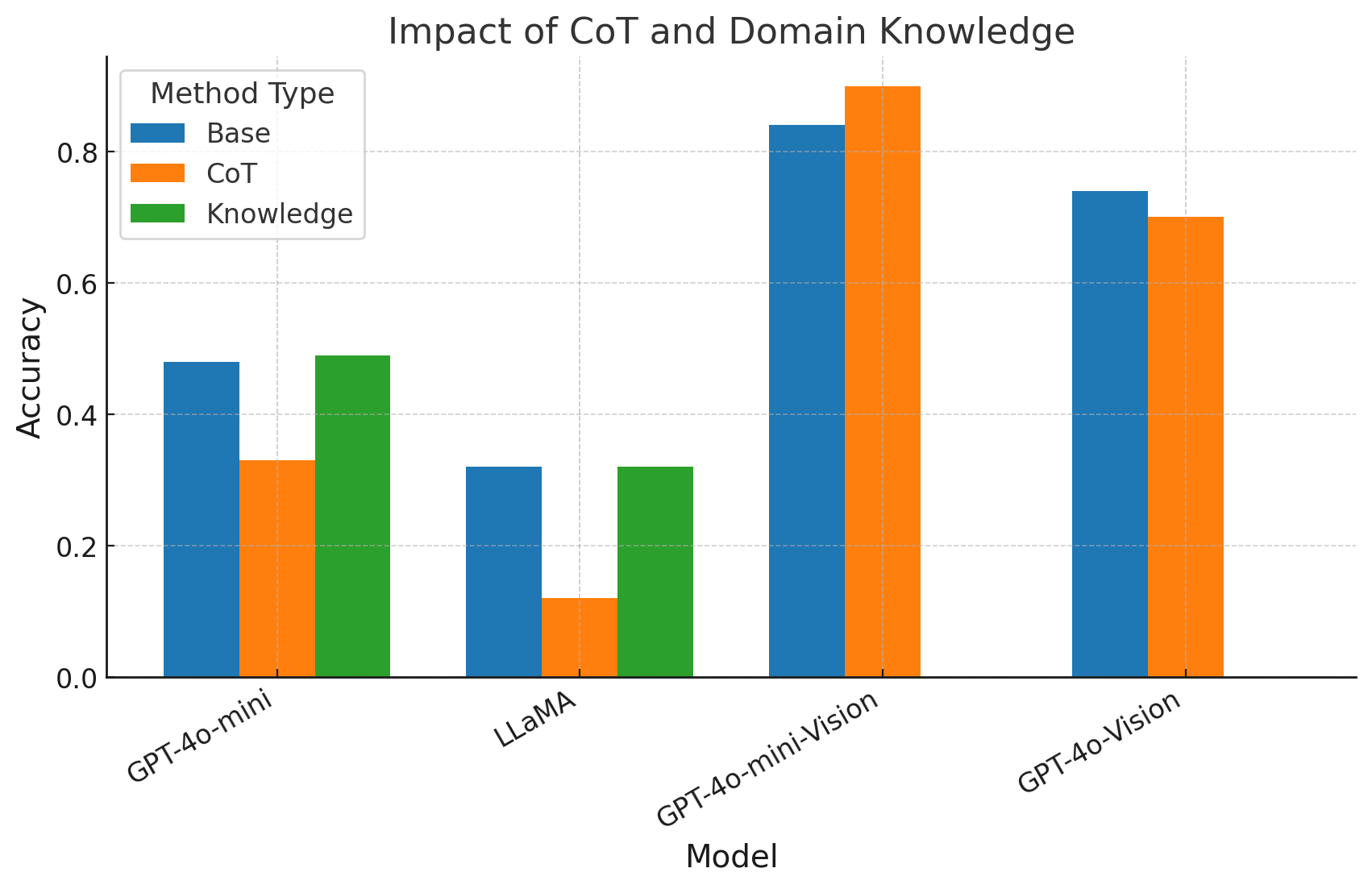}
    \caption{Comparison of COT and knowledge}
    \label{bar}
\end{figure}

\textbf{Effectiveness of CoT and Domain Knowledge.}
Figure~\ref{bar} illustrates the impact of CoT reasoning and domain knowledge on different language models across time series and vision-based settings. Among the zero-shot models, GPT-4o-mini demonstrates notable improvements when domain knowledge is incorporated, highlighting the benefits of integrating prior knowledge into the inference process. However, a substantial performance drop is observed in both models when CoT reasoning is applied to time series data. We argue that this decline stems from the inherent complexity of time series signals, particularly in the case of Wi-Fi raw signals, which are challenging to interpret directly. The step-by-step reasoning introduced by CoT may inadvertently add confusion, as the model struggles to generate coherent intermediate steps for highly dynamic and noisy input sequences.  

For vision-enhanced models, the GPT-4o-mini-Vision variant achieves the highest performance, with CoT prompting yielding an accuracy of 0.90. This suggests that CoT reasoning is particularly effective when paired with visual input, likely because images provide additional context that facilitates structured reasoning. A similar, albeit less pronounced, effect is observed for GPT-4o-Vision. These results indicate that while CoT reasoning can be beneficial, its effectiveness varies depending on the specific vision-language model and its alignment with the task. The structured nature of visual input may better support multi-step reasoning, whereas time series data lacks the same level of interpretability, limiting the effectiveness of CoT in those scenarios.  

\begin{figure}[ht]
    \centering
    \includegraphics[width=0.49\textwidth]{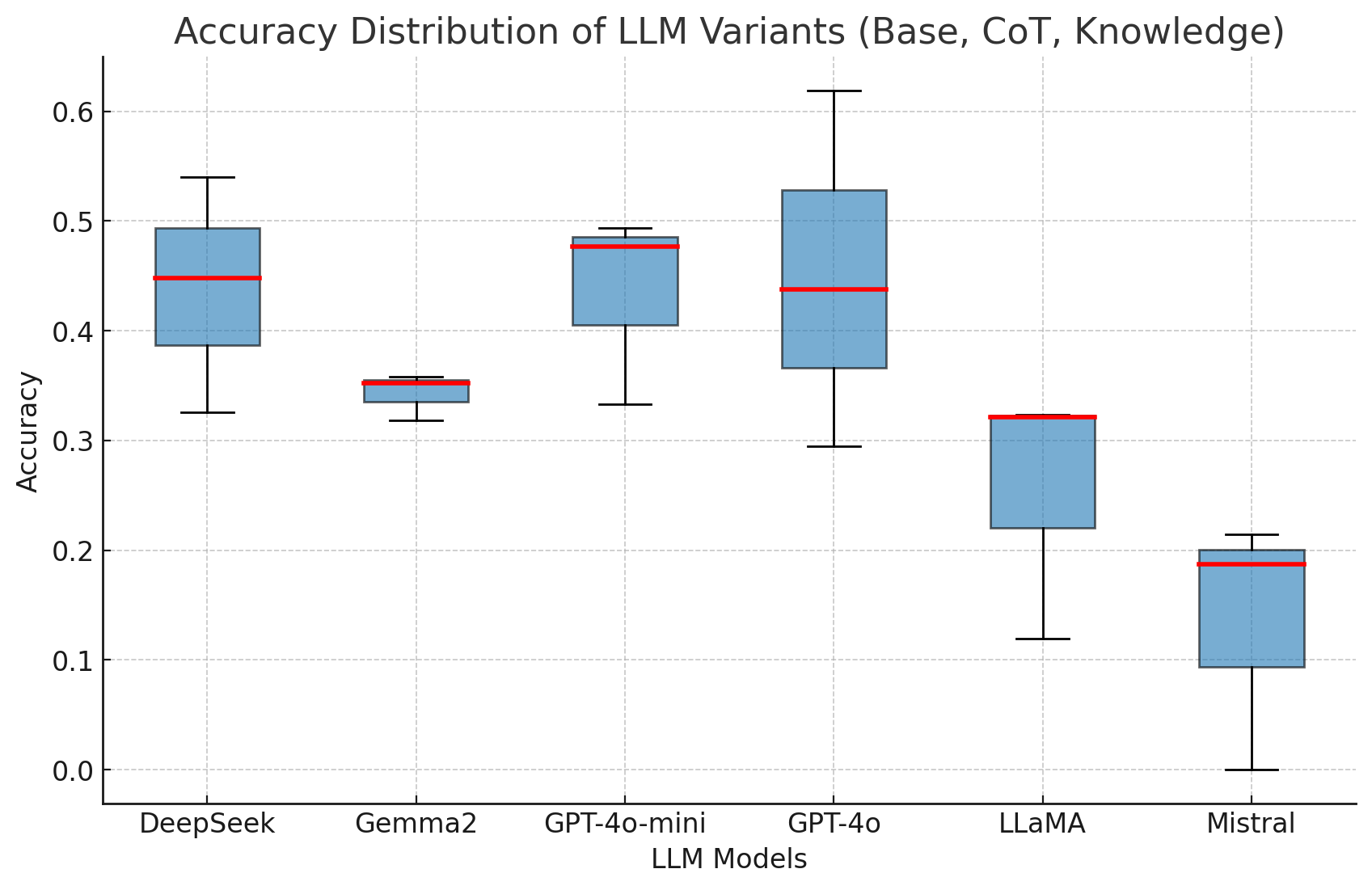}
    \caption{Comparison of LLMs under zero-shot settings}
    \label{box}
\end{figure}

\textbf{Comparing Different LLMs.} Figure~\ref{box} presents the accuracy distribution of different LLM variants (Base, CoT, and Knowledge) under zero-shot settings. The box plots illustrate the variability in accuracy across models, with GPT-4o and DeepSeek demonstrating the highest median performance, while Mistral and LLaMA exhibit greater variance and lower median accuracy.

Among the models, GPT-4o-mini and DeepSeek show a more stable accuracy distribution, indicating consistent performance across different variants. In contrast, LLaMA has a wider spread, suggesting that its performance is more sensitive to the specific reasoning approach used. Gemma2 maintains a relatively narrow distribution, indicating lower variance but also a more limited improvement potential.

These results highlight that larger models like GPT-4o benefiting from more robust reasoning capabilities, while smaller models show varying degrees of improvement depending on the applied enhancements.


\section{Conclusion}
In this paper, we introduced Wi-Chat, the first LLM-powered Wi-Fi-based human activity recognition system that integrates the reasoning capabilities of large language models with the sensing potential of wireless signals. Our experimental results on a self-collected Wi-Fi CSI dataset demonstrate the promising potential of LLMs in enabling zero-shot Wi-Fi sensing. These findings suggest a new paradigm for human activity recognition that does not rely on extensive labeled data. We hope future research will build upon this direction, further exploring the applications of LLMs in signal processing domains such as IoT, mobile sensing, and radar-based systems.

\section*{Limitations}
While our work represents the first attempt to leverage LLMs for processing Wi-Fi signals, it is a preliminary study focused on a relatively simple task: Wi-Fi-based human activity recognition. This choice allows us to explore the feasibility of LLMs in wireless sensing but also comes with certain limitations.

Our approach primarily evaluates zero-shot performance, which, while promising, may still lag behind traditional supervised learning methods in highly complex or fine-grained recognition tasks. Besides, our study is limited to a controlled environment with a self-collected dataset, and the generalizability of LLMs to diverse real-world scenarios with varying Wi-Fi conditions, environmental interference, and device heterogeneity remains an open question.

Additionally, we have yet to explore the full potential of LLMs in more advanced Wi-Fi sensing applications, such as fine-grained gesture recognition, occupancy detection, and passive health monitoring. Future work should investigate the scalability of LLM-based approaches, their robustness to domain shifts, and their integration with multimodal sensing techniques in broader IoT applications.

\bibliography{main}
\newpage
\appendix

\section{Experiment prompts}
\label{sec:prompt}
The prompts used in the LLM experiments are shown in the following Table~\ref{tab:prompts}.
\begin{table}[ht]
    \centering
    \small
    \caption{Examples of Prompts Used in Experiments}
    \begin{tabular}{p{3cm} p{10cm}}
        \toprule
        \textbf{Prompt Type} & \textbf{Example Prompt} \\
        \midrule
        \textbf{Zero-shot} & 
        ``You are given a time series of Channel State Information (CSI) amplitude values captured in an environment. Your task is to classify the activity into one of the following four categories: breath, fall, no event, or walk. The provided data represents the amplitude variations over time. Analyze the pattern and determine the most appropriate classification label based on the observed fluctuations. \newline
        \textbf{Input Data:} \{time\_series\_data\} \newline
        No need to explain the reason. Output only one word as the predicted activity label: breath, fall, no event, or walk.'' \\
        \midrule
        \textbf{Knowledge} & 
        ``You are an expert in Channel State Information (CSI)-based human activity recognition. CSI data reflects the changes in wireless signals as they interact with human movements. Based on CSI amplitude variations, human activities can be categorized into four types:\newline
        \textbf{Walking:} Large and multiple changes in amplitude over time. \newline
        \textbf{Falling:} A single significant peak or trough followed by a relatively stable period. \newline
        \textbf{Breathing:} Smooth and moderate changes in amplitude over time. \newline
        \textbf{No-event:} The amplitude remains mostly stable with minimal fluctuations. \newline
        \textbf{Input Data:} \{time\_series\_data\} \newline
        Your task is to classify the following CSI time series into one of these four categories. \newline
        No need to explain the reason. Output only one word as the predicted activity label: breath, fall, no event, or walk.'' \\
        \midrule
        \textbf{Chain-of-Thought (CoT)} & 
        ``You are an expert in Channel State Information (CSI)-based human activity recognition. CSI data reflects the changes in wireless signals as they interact with human movements. Human bodies, being significant obstacles, cause changes in wireless signals that CSI can capture.\newline
        You will receive time series CSI amplitude data recorded when a person is performing an activity. The person's activity belongs to one of the following categories: Walking, Falling, Breathing, or No-event.\newline
        \textbf{Step-by-Step Analysis of Human Activity:}\newline
        - What is the variation range of the data? If the variation range is very small, does it suggest a no-event scenario?\newline
        - Does the data change smoothly over time with a moderate variation range (e.g., less than 5)? If so, could this indicate breathing?\newline
        - Identify large peaks or troughs characterized by a significant increase or decrease, which then return to the overall range.\newline
        - How many large peaks or troughs are present in the data?\newline
        - If the data contains only one significant peak or trough (dramatically larger than others), does it transition into a relatively stable period with minor variations? If so, does this indicate a falling event?\newline
        - If there are multiple large peaks and troughs occurring regularly, does this suggest a walking activity?\newline
        \textbf{Final Classification:} Based on the above step-by-step analysis, determine the most appropriate classification for the given time series data.\newline
        \textbf{Input Data:} \{time\_series\_data\}\newline
        No need to explain the reason. Output only one word as the predicted activity label: breath, fall, no event, or walk.'' \\
        \bottomrule
    \end{tabular}
    \label{tab:prompts}
\end{table}


\end{document}